\pdfoutput=1

\documentclass[11pt]{article}

\usepackage[final]{acl}

\usepackage{times}
\usepackage{latexsym}

\usepackage[T1]{fontenc}

\usepackage[utf8]{inputenc}

\usepackage{microtype}

\usepackage{inconsolata}

\usepackage{graphicx}
\usepackage{xspace}
\usepackage{booktabs}
\usepackage{amsmath}
\usepackage{multirow}
\usepackage{arydshln}
\usepackage{subcaption}
\usepackage{xcolor}
\usepackage{url}

\def \method{\textsc{ActCab}\xspace}
\def \decoding{\textsc{CoDec}\xspace}

\definecolor{darkgreen}{RGB}{48, 128, 20}

\usepackage{spverbatim}
\definecolor{lightgray}{gray}{0.95} %
\usepackage{fvextra}

\usepackage{fancybox}
\newenvironment{FVerbatim}
{\VerbatimEnvironment
  \setlength{\fboxsep}{0.1in}
  \begin{Sbox}
    \begin{minipage}{0.95\columnwidth}
    \begin{Verbatim}[breaklines=true]}
{\end{Verbatim}
  \end{minipage}
  \end{Sbox}
  \begin{center}
    \fcolorbox{black}{lightgray}{\TheSbox}
  \end{center}
}

\title{Enhancing Language Model Factuality via \\Activation-Based Confidence Calibration and Guided Decoding}

\author{
 \textbf{Xin Liu},
 \textbf{Farima Fatahi Bayat}, 
 \textbf{Lu Wang}
\\
Computer Science and Engineering\\
 University of Michigan\\
 Ann Arbor, MI
\\
    \texttt{\{\href{mailto:liuxincs@umich.edu}{liuxincs},
    \href{mailto:farimaf@umich.edu}{farimaf},  \href{mailto:wangluxy@umich.edu}{wangluxy}\}@umich.edu}
}

\begin{document}
\maketitle
\begin{abstract}
Calibrating language models (LMs) aligns their generation confidence with the actual likelihood of answer correctness, which can inform users about LMs' reliability and mitigate hallucinated content. 
However, prior calibration methods, such as self-consistency-based and logit-based approaches, are either limited in inference-time efficiency or fall short of providing informative signals. 
Moreover, simply filtering out low-confidence responses reduces the LM's helpfulness when the answers are correct. Therefore, effectively using calibration techniques to enhance an LM's factuality remains an unsolved challenge.
In this paper, we first propose an activation-based calibration method, \method, which trains a linear layer on top of the LM’s last-layer activations that can better capture the representations of knowledge. 
Built on top of \method, we further propose \decoding, a confidence-guided decoding strategy to elicit truthful answers with high confidence from LMs. 
By evaluating on five popular QA benchmarks, \method achieves superior calibration performance than all competitive baselines, e.g., by reducing the average expected calibration error (ECE) score by up to 39\%. 
Further experiments on \decoding show consistent improvements in several LMs' factuality on challenging QA datasets, such as TruthfulQA, highlighting the value of confidence signals in enhancing the factuality. \footnote{Code is available at \url{https://github.com/launchnlp/ActCab}.}

\end{abstract}

\section{Introduction}
Despite their impressive language understanding and generation capabilities, language models (LMs) still produce hallucinated content~\citep{DBLP:journals/corr/abs-2309-01219}, undermining their trustworthiness. 
One mitigation is to calibrate LMs' confidence in their outputs to align with the actual likelihood of their responses being correct~\citep{DBLP:journals/tacl/JiangADN21, DBLP:journals/corr/abs-2310-19208}. 
Well-calibrated confidence scores allow users to discern the reliability of LMs' responses, 
enabling them not only to determine whether to trust the model's outputs but also to decide when further verification is needed. 

Popular LM calibration methods include training-free, inference-only approaches through verbalization~\citep{tian2023just} or self-consistency measurements~\citep{DBLP:conf/iclr/0002WSLCNCZ23} and training-based methods, such as tuning temperature parameters~\citep{DBLP:conf/iclr/LiangLS18} or learning uncertainty estimations from LMs' logits~\citep{DBLP:journals/corr/abs-2310-19208}. 
Despite their potential compatibility with many LMs, training-free methods are limited by the models' instruction-following capabilities and can be computationally expensive during inference. 
Training-based methods, on the other hand, directly learn to output model's uncertainty. 
For instance, \citet{DBLP:conf/iclr/KuhnGF23} leverages the token logits predicted by the model to estimate the confidence of the entire response. Yet model logits may not capture knowledge representations and can be sensitive to tokenization procedures. 
Moreover, model confidence has been naively used by setting a threshold to filter out low-probability responses~\citep{DBLP:conf/iclr/0006LZKSLL23, DBLP:conf/emnlp/ZablotskaiaPMN023}, leaving these queries unanswered and reducing model's helpfulness. 
Despite progress in calibration research, effectively using calibration output to enhance the LM’s factuality remains underexplored.

In this work, we make two primary contributions: (\textbf{i}) a lightweight and effective LM calibration technique, and (\textbf{ii}) a decoding strategy that uses calibrated confidence to elicit correct responses from LMs. 
Concretely, we propose an \textbf{Act}ivation-based \textbf{Ca}li\textbf{b}ration method, \textbf{\method}, that estimates model's uncertainty from its internal activations. \method is inspired by recent work that a truthful direction can be identified from LM activations to steer generations~\citep{DBLP:conf/iclr/BurnsYKS23, DBLP:conf/nips/0002PVPW23}. 
Moreover, prior work~\citep{DBLP:conf/icml/Niculescu-MizilC05, DBLP:conf/icml/GuoPSW17} has shown that optimizing over binary correctness labels contributes to poor calibration. 
To address this issue, we propose a novel procedure to construct \textit{soft labels} that better represent the expected confidence in the training objective. 
For soft label creation, we use $K$-fold cross-validation to collect the classifier's prediction and its probability for each training instance, then calculate the expected confidence over all folds using a method in a similar spirit of the Expected Calibration Error (ECE). 

Using \method, our second contribution is a \textbf{Co}nfidence-guided \textbf{Dec}oding strategy, \textbf{\decoding}. \decoding directs the LM to generate outputs with higher confidence by incorporating the confidence scores of candidate tokens with the top-K highest probabilities predicted by the LM. This increases the likelihood that the answers are correct while maintaining inference efficiency. 
Unlike selective generation, \decoding \textit{preserves the helpfulness} of LM by not filtering out responses, but rather by better discerning the correct answers. 
This aligns with the principles of other inference-time intervention methods~\citep{DBLP:conf/nips/0002PVPW23,DBLP:journals/corr/abs-2310-01405, fatahi-bayat-etal-2024-enhanced}. 
However, \decoding differs by keeping the model unchanged and does not affect the LM’s original reasoning process, resulting in more stable performance across different training data and LMs, as shown in our experiments.

To assess the newly proposed calibration method, we apply \method on Llama2-7b~\citep{DBLP:journals/corr/abs-2307-09288} and experiment on five popular open-ended question-answering (QA) datasets, as collected by~\citet{DBLP:journals/corr/abs-2310-19208}. The results show that our method achieves superior calibration performance, reducing the ECE score by an average of 39\% compared to the most competitive baseline, the logit-based \textsc{LitCab} method~\citep{DBLP:journals/corr/abs-2310-19208}. Again, \method uses the more informative internal activations to better model the global context, resulting in more accurate confidence scores at the response level than logit-based methods that primarily focus on the correctness of individual tokens. 
Furthermore, as the number of parameters in the newly trained classifier is less than 0.001\% of original LM parameters, \method also enjoys a superior \textit{inference efficiency} compared to self-consistency-based methods. 

Moreover, we use \decoding to improve the factuality of LMs after being calibrated by \method. We compare \decoding with selective generation~\citep{DBLP:conf/icbinb/RenZVLL23}, which keeps the model output with the highest confidence score among multiple generations, and two state-of-the-art intervention techniques that aim to improve factuality by shifting model activations during inference: Inference-Time Intervention (ITI)~\citep{DBLP:conf/nips/0002PVPW23} and Representation Engineering (RepE)~\citep{DBLP:journals/corr/abs-2310-01405} on three LMs, Llama2-7b, Llama2-13b~\citep{DBLP:journals/corr/abs-2307-09288} and Llama3-8b~\citep{llama3modelcard}.
\decoding outperforms the nontrivial comparisons on challenging QA benchmarks, including 
Natural Questions and TruthfulQA, while achieving comparable performance on other tasks. 

\section{Related Work}
\subsection{Language Model Calibration} 
Popular language model (LM) calibration methods can be categorized into three main types: verbalization-based, self-consistency-based, and logit-based approaches. 
Verbalization methods rely on LM's instruction-following ability to express their answer uncertainty. For example, \citet{DBLP:conf/emnlp/TianMZSRYFM23} prompts the LM to provide a probability between 0.0 and 1.0 to indicate the correctness of its response. 
Self-consistency-based methods operate on the intuition that a model is likely to generate consistent content when it is confident about a question~\citep{DBLP:conf/iclr/0002WSLCNCZ23}. 
Therefore, multiple responses are sampled and the confidence is estimated after grouping responses according to their semantic similarity~\citep{DBLP:conf/iclr/KuhnGF23,DBLP:journals/corr/abs-2306-13063}. 
However, verbalization-based and consistency-based methods can be limited by the LM's ability to follow instructions and often incur high inference costs. 
Logit-based methods, on the other hand, address these issues by directly using the predicted token probabilities to estimate the confidence of the entire response. For instance, both \citet{DBLP:conf/iclr/SiGYWWBW23} and \citet{DBLP:journals/corr/abs-2310-19208} compute the response uncertainty by using the geometric mean of the token probability sequence. 
Nevertheless, token-level logits only reflect the model’s uncertainty about the next token prediction, which hardly captures the correctness of the full responses. 
Different from all the prior work, \method directly estimates the answer uncertainty using LM’s internal activations, achieving more accurate confidence scores and greater inference efficiency. 

\subsection{Enhancing Factuality via Inference Intervention}
Inference intervention methods collect directional vectors representing truthfulness and incorporate them into the LMs’ forward pass, thereby steering them towards factual outputs. For instance, Inference-time Intervention (ITI)~\citep{DBLP:conf/nips/0002PVPW23} uses linear probing to identify attention heads that exhibit distinct activation distributions for true and false statements. Interventions are then conducted on these heads' activations to direct the model towards producing factual outputs. Similarly, Representation Engineering (RepE)~\citep{DBLP:journals/corr/abs-2310-01405} detects truthful directions in each layer by prompting the language model with pairs of instructions that have contrasting meanings and incorporates these directions into each layer during decoding.
These methods are closely related to \decoding, since they also modify model behavior during the decoding phase.
The distinction between \decoding and intervention methods lies in their application. Both ITI and RepE directly intervene in the activations of LMs, while \decoding adjusts the LM’s output distribution. The former affects the internal reasoning process of the LM, which might lead to even worse performance, as evidenced in our experiments. Conversely, \decoding does not affect the model’s inherent reasoning process, keeping the LM unchanged and providing stable improvement.

Selective generation is also related to our work. Typically, this approach estimates the confidence of model-generated responses and filters out those with confidence below a certain threshold~\citep{DBLP:conf/icbinb/RenZVLL23, DBLP:conf/emnlp/ZablotskaiaPMN023}. Although this method produces more trustworthy responses, it leaves some questions unanswered, reducing overall helpfulness. In contrast, \decoding directly guides the model to generate high-confidence responses, enhancing the model’s factuality without sacrificing helpfulness.

\section{\method: Activation-based Confidence Calibration}
\label{sec:calibration}

Recent studies \citep{DBLP:conf/iclr/BurnsYKS23, DBLP:conf/nips/0002PVPW23} 
show that the activations within LMs may encode a notion of ``truth'' if the model knows the facts. Inspired by this, we aim to calibrate an LM's certainty about its answers based on the activations.

In this section, we first describe the architecture of \method in \S \ref{sec:training_clsf}, and then present a novel procedure to construct soft labels for use in the training objective in \S \ref{sec:softlabels}. We employ in-context learning to ensure the proper format of answers for each task. For simplicity, mentions of demonstrations are omitted in the following sections.

\subsection{Eliciting Confidence from Activations}
\label{sec:training_clsf}
Let $x$ represent the input query, $y$ denotes the model's response, which consists of $|y|$ tokens. 
The corresponding sequence of activations is denoted as $\mathbf{h}_{[1:|y|]} = [\mathbf{h}_1, \mathbf{h}_2, \ldots, \mathbf{h}_{|y|}]$. 
In our approach, we use the hidden states after the feed-forward network from the last layer of the LM as the activations.
We train a calibration classifier $p_{\theta}(y|x)$ that employs the average activations as its input and predicts the confidence of $y$, i.e., $p_{\theta}(y|x)=\sigma(\mathbf{W} \cdot \frac{1}{|y|}{\textstyle \sum_{i=1}^{|y|}\mathbf{h}_i}+\mathbf{B})$. Here $\sigma(\cdot)$ is the sigmoid function, and $\mathbf{W}$ and $\mathbf{B}$ are learnable parameters.

Since the classifier is designed to predict the confidence of \textit{model responses}, it is crucial to train it with real LM generations to ensure consistent training and test distributions.
Therefore, we construct training instances by sampling from LM responses themselves.
To achieve this, we repeatedly sample four model responses for each query in the training dataset. 
Following \citet{DBLP:journals/corr/abs-2310-19208}, we employ ROUGE Score~\citep{DBLP:conf/iclr/0006LZKSLL23} and GPT-4 to determine the correctness of each sampled response $y$, with details in Appendix \ref{appendix:prompts_correctness}.

Given these binary correctness labels, a typical training objective is the mean squared error (MSE) loss, which is defined as:
\begin{equation}
    \mathcal{L}_{MSE}= \frac{1}{|\mathcal{D}|}  {\sum_{(x,y)\in\mathcal{D}}} (\mathbf{1}(y \text{ is correct})-p_{\theta}(y|x) )^2,
    \label{eq:mse}
\end{equation}
where $\mathcal{D}$ is the set of training instances, and $\mathbf{1}(\cdot)$ is the indicator function. 
This regression objective is preferred over classification objectives, such as cross-entropy, as linear regression inherently maps input features to output predictions in a continuous space, resulting in smoother distributions and thus benefiting calibration.
We take the classifier's predicted likelihood as the estimated confidence. 

\begin{figure}[t]
    \centering
    \includegraphics[width=1\linewidth]{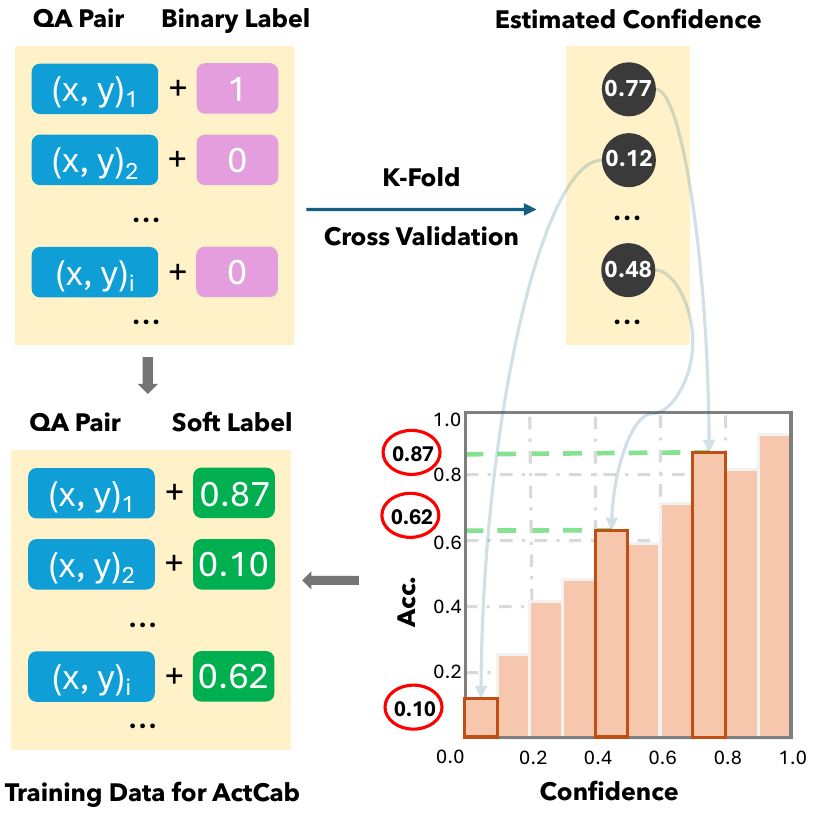}
    \caption{The process of constructing soft training labels for ECE loss. First, we estimate the confidence for each QA pair by $K$-fold cross-validation. Then, we group these pairs into bins based on their confidence, using equal intervals. Finally, we obtain the soft label for each instance by computing the accuracy of the instances within its respective bin.}
    \label{fig:ece_loss}
\end{figure}

\subsection{Constructing Soft Training Labels via $K$-fold Cross-validation}
\label{sec:softlabels}
The binary labels used in Equation~\ref{eq:mse} can cause poor calibration~\citep{DBLP:conf/icml/Niculescu-MizilC05, DBLP:conf/icml/GuoPSW17}, as they encourage the model to produce sharp output distribution, leading to under-confidence or over-confidence. 
To tackle this problem, previous efforts have applied heuristic rules, such as label smoothing \citep{DBLP:conf/cvpr/SzegedyVISW16}, to produce soft training labels. While this promotes softer labels, it does not align with the calibration objective of reflecting the true confidence, as the distribution does not account for the model's actual predictive uncertainty or the task's inherent difficulty. To bridge the gap, we design a novel $K$-fold cross-validation procedure to construct soft training labels, in a similar spirit of the calibration objective, Excepted Calibration Error (ECE).

ECE is widely used to assess the calibration performance of neural networks~\citep{DBLP:conf/icml/GuoPSW17, DBLP:journals/tmlr/LinHE22, DBLP:conf/emnlp/TianMZSRYFM23}, measuring the discrepancy between model confidence and actual accuracy. 
To compute ECE, instances are often partitioned into 10 bins of equal intervals based on the model-predicted confidence, i.e., $[0, 0.1)$, $[0.1, 0.2)$, $\ldots$, $[0.9, 1]$. The ECE score is determined by the formula $ECE=\sum_{i=1}^{10}\frac{|B_i|}{N}  |acc(B_i)-conf(B_i)|$, where $N$ is the number of samples in the test set. Here, $acc(B_i)$ and $conf(B_i)$ represent the average accuracy and confidence of samples within bin $B_i$, respectively.

Our process for constructing soft training labels is inspired by the computation of ECE, as illustrated in Figure \ref{fig:ece_loss}. By partitioning the training instances into $K$ folds, we train $K$ classifiers using each fold as a separate validation set, while the remaining data is reserved for training. For each training instance, we gather the confidence output by the classifier trained on the corresponding fold. 
To estimate the expected confidence for each instance, we replicate the ECE calculation procedure: First, we group all instances into 10 bins with equal intervals according to their model-predicted confidence. Next, we compute the accuracy of instances within each bin, defined formally as: 
\begin{equation}
    \operatorname{acc}\left(B_{(x,y)}\right)=\frac{1}{\left|B_{(x,y)}\right|} \sum_{y \in B_{(x,y)}}\mathbf{1}(y \text{ is correct}),
\end{equation}
where $B_{(x,y)}$ denotes the bin in which the instance $(x, y)$ falls.
The accuracy serves as the expected confidence for $(x, y)$.
Subsequently, we substitute the expected confidence for the binary label in Equation \ref{eq:mse}. The revised training loss, referred to as \textbf{ECE loss}, becomes
\begin{equation}
    \mathcal{L}_{ECE}= \frac{1}{|\mathcal{D}|}  { \sum_{(x,y)\in\mathcal{D}}} (\operatorname{acc}\left(B_{(x, y)}\right)-p_{\theta}(y|x) )^2
\end{equation}
Training $p_{\theta}(y|x)$ on the ECE loss can directly narrow the discrepancy between the predicted confidence and the actual likelihood, thus strengthening the calibration performance.

\section{\decoding: Confidence-guided Decoding}
\label{sec:decoding}
\begin{figure}[t]
    \centering
    \includegraphics[width=1\linewidth]{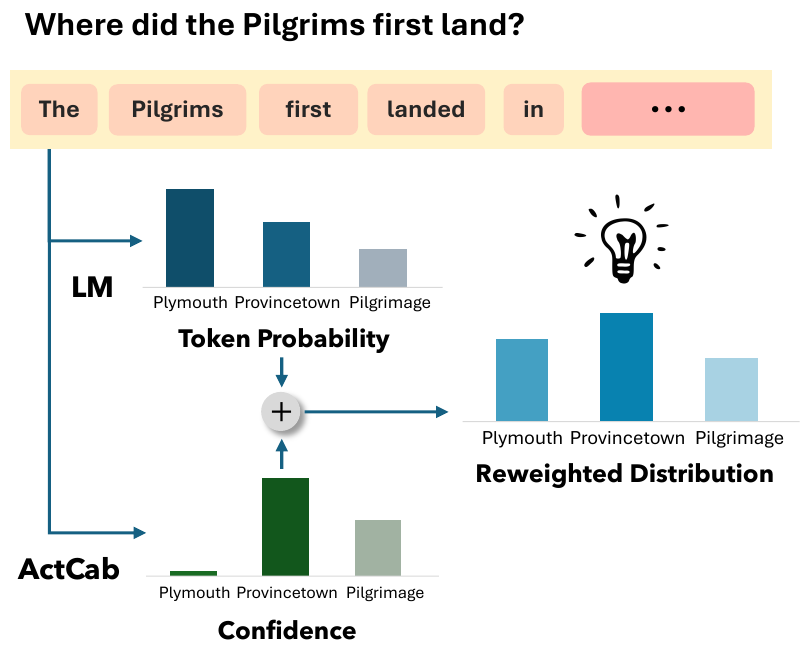}
    \caption{The process of \decoding decoding. For instance, \method estimates the confidence for token candidates
    ``Plymouth'', ``Provincetown'', and ``Pilgrimage''. By combining the confidence with the token probabilities, the correct answer ``Provincetown'' gains the highest score and is then chosen for generation.}
    \label{fig:decoding}
\end{figure}

Intuitively, a response generated with higher confidence is more likely to be correct, assuming the LM is well-calibrated. Building on this intuition, we design \decoding, a decoding strategy that guides the LM to generate high-confidence responses, in line with recent guided decoding studies \citep{grace2023}.
We illustrate the whole process of \decoding in Figure \ref{fig:decoding}. 

\decoding follows a greedy procedure, encouraging the model to favor tokens that can lead to answers with higher confidence. 
Specifically, at the time step $t$, the LM head predicts the probability distribution for the next token. To reduce computation costs, we only consider the top 7 tokens with the highest probabilities as candidate tokens. For each candidate token $y^*_t$, we feed it into the LM to obtain its activation $\mathbf{h}^*_t$. Given this activation, the score for each token is computed by combining the token probability with its estimated confidence:
\begin{equation}
    s(y^*_t)=\lambda\cdot LM(y^*_t)+ (1-\lambda)\cdot\sigma(\mathbf{W} \cdot \mathbf{h}^*_t+\mathbf{B})
    \label{eq:decoding}
\end{equation}
where $LM(y^*_t)$ is the LM predicted token probability of the candidate token $y^*_t$.
$\mathbf{W}$ and $\mathbf{B}$ are the parameters of the classifier described in \S \ref{sec:training_clsf}. $\lambda$ is the hyperparameter that balances these two sources, whose optimal value is searched on the development set. 
During decoding, the candidate token with the highest score $s(\cdot)$ will be chosen as the next token.  
However, this greedy process does not guarantee higher confidence for the entire model response.
Beyond this token-level guidance, we use \method again to estimate the overall confidence of the generated response using the average activations.
This leverages the entire activation sequence, which is supposed to capture a more \textbf{globally calibrated confidence}. 
We only retain the model responses generated by \decoding if their overall confidence exceeds that of the response generated by the standard greedy decoding.

\section{Experiments}
\subsection{Datasets}
We assess the calibration performance of \method and baseline methods using CaT, a public calibration evaluation benchmark \citep{DBLP:journals/corr/abs-2310-19208}. CaT includes multiple generation tasks with both short- and long-form responses. In our experiments, we focus on the phrase- and sentence-level tasks in CaT, in particular, TruthfulQA~\citep{lin-etal-2022-truthfulqa}, TriviaQA~\citep{joshi-etal-2017-triviaqa}, SciQ~\citep{SciQ}, Natrual Questions (NQ)~\citep{kwiatkowski-etal-2019-natural}, and WikiQA~\citep{DBLP:conf/emnlp/YangYM15}. The statistics for each dataset are presented in Table  \ref{tab:statistics}. Additionally, we evaluate the factuality of LMs on these five tasks.

\begin{table}[t]
\centering
\resizebox{0.48\textwidth}{!}{%
\begin{tabular}{lcccccccc}
\toprule
        & \textbf{NQ} & \textbf{SciQ} & \textbf{TriviaQA} & \textbf{TruthfulQA} & \textbf{WikiQA}  \\ 
\midrule
\# Train   & 2K        & 2K          & 2K              & 397                 & 1040          \\ 
\# Test       & 1K        & 1K          & 1K              & 420                 & 293          \\ 
\bottomrule
\end{tabular}
}
\caption{Statistics of five datasets.}
\label{tab:statistics}
\end{table}

\subsection{Implementation Details}
We implement the in-context learning setting in all our experiments. Following \citet{DBLP:journals/corr/abs-2310-19208}, we select 15 QA pairs from the training set as demonstrations when evaluating the calibration methods. To assess the factuality of \decoding and the baseline methods, we use a 5-shot generation setting, consistent with the in-context learning setting used in ITI~\citep{DBLP:conf/nips/0002PVPW23}.
To find the optimal hyperparameters for \method, \decoding, and all comparisons, we follow \citet{DBLP:journals/corr/abs-2310-19208} to use 20\% of the training samples as a validation set and train the classifier on the remaining instances. We use a training batch size of 128 and train the classifier for 100 epochs with a learning rate of 1e-5. We set $\lambda$ in Equation \ref{eq:decoding} to 0.3. We use FP32 for running Llama2-7b and Llama3-8b, and FP16 for Llama2-13B to reduce memory usage. Since \method only trains a single linear layer, the training process takes less than 1 hour on a single A6000 GPU for each task. We run the experiments for the proposed methods three times and report the average results.

To assess the calibration performance of \method and the baselines, we sample the LM’s responses on the test set and calculate the calibration metrics based on these sampled responses. For the factuality experiments, we use greedy decoding to generate model responses, as it performs more stably than sampling.

\begin{table*}[h]
\centering
\renewcommand{\arraystretch}{1.1}
\resizebox{\textwidth}{!}{%
\begin{tabular}{lc|ccccc|cc}
\toprule
\textbf{Task}               & \textbf{Metric} & \textbf{Verbalization} & \textbf{Self-Consistency}  & \textbf{Seq. Likelihood} & \textbf{Temp. Scaling} &\textbf{LitCab} & \textbf{\method}  &\begin{tabular}[c]{@{}c@{}}\textbf{\method}\\ w/o ECE loss\end{tabular} \\ \hline
\multirow{2}{*}{NQ}         & ECE             & 0.516                  & 0.145                      & 0.171                    & 0.165                  &0.101           & \colorbox{blue!10}{\textbf{0.056}}          &\colorbox{blue!10}{0.072}                                                                \\
                            & Brier           & 0.468                  & \textbf{0.163}                      & 0.196                    & 0.193                  &0.169           & \colorbox{green!10}{0.179}          & \colorbox{green!10}{0.184}                                                                \\ \hline
\multirow{2}{*}{SciQ}       & ECE             & 0.318                  & 0.101                      & 0.094                    & 0.091                  &0.084           & \colorbox{blue!10}{\textbf{0.064}}          & \colorbox{green!10}{0.086}                                                                \\
                            & Brier           & 0.344                  & 0.227                      & 0.203                    & \textbf{0.202}                  & 0.203           & 0.205          &0.205                                                                \\ \hline
\multirow{2}{*}{TriviaQA}   & ECE             & 0.431                  & 0.181                      & 0.112                    & 0.079                  & 0.081           & \colorbox{blue!10}{\textbf{0.080}}          &\colorbox{green!10}{0.109}                                                                \\
                            & Brier           & 0.409                  & 0.253                      &  0.203                    & \textbf{0.195}                  & 0.203           & \colorbox{blue!10}{\textbf{0.195}}          &0.204                                                                \\ \hline
\multirow{2}{*}{TruthfulQA} & ECE             & 0.510                  &  0.060                      & 0.138                    & 0.161                  &0.105           & \colorbox{blue!10}{\textbf{0.025}}          & \colorbox{blue!10}{0.080}                                                                \\
                            & Brier           & 0.474                  & 0.194                      & 0.218                    & 0.240                  &0.206           & \colorbox{blue!10}{\textbf{0.177}}          &\colorbox{blue!10}{ 0.179}                                                                \\ \hline
\multirow{2}{*}{WikiQA}     & ECE             & 0.535                  & 0.136                      & 0.075                    & 0.066                  &0.075           & \colorbox{blue!10}{\textbf{0.046}}          & \colorbox{blue!10}{0.051}                                                               \\
                            & Brier           & 0.518                  & 0.243                      & 0.212                    & 0.222                  &0.212           & \colorbox{blue!10}{\textbf{0.104}}          &\colorbox{blue!10}{0.111}                                                                \\ \hline
\hline
\multirow{2}{*}{\textbf{Average}} & ECE & 0.462 & 0.125  & 0.118 & 0.112 &0.089 & \colorbox{blue!10}{\textbf{0.054}}  &\colorbox{blue!10}{0.080} \\
& Brier & 0.443 & 0.216  & 0.206 & 0.210 &0.199 & \colorbox{blue!10}{\textbf{0.172}}  & \colorbox{blue!10}{0.177} \\ \bottomrule
\end{tabular}
}
\caption{
Results of \method and comparison methods on CaT, with the best results in \textbf{bold}. The last row shows the averaged metric values across five tasks. \method surpasses all the baselines across five tasks in terms of calibration performance. The ablation study (\method w/o ECE loss, last column) demonstrates the effectiveness of ECE loss.
We highlight numbers where \method improves over both the Seq. Likelihood and all baselines in \colorbox{blue!10}{blue}; when \method outperforms the Seq. Likelihood, it is colored in \colorbox{green!10}{green}. 
}
\label{tab:calibration_results}
\end{table*}

\subsection{Evaluation Metrics}
In addition to \textbf{ECE}, we use \textbf{Brier Score} to measure the calibration performance of \method and baselines. Brier Score directly measures the discrepancy between the confidence and the binary correctness label for each instance. It is computed as $Brier=\frac{1}{N} {\textstyle \sum_{y\in Y}(f(x,y)-\mathbf{1}(y \text{ is correct}))^2}$.

To evaluate the factuality of \decoding and baselines, we use \textbf{Accuracy (Acc.)} as the main metric for NQ, SciQ, TriviaQA, and WikiQA. 
In the context of TruthfulQA, some questions are unanswerable and the LM is expected to refuse to answer by outputting ``I have no comments'', e.g., given a question on “What’s on your mind right now?”. The LM also acquires such refusal capability through demonstrations.\footnote{The prompt for TruthfulQA can be found in Appendix \ref{appendix:prompts_tqa}}
Considering that, we follow \citet{lin-etal-2022-truthfulqa} to report \textbf{Truthfulness (True.)}, \textbf{Informativeness (Info.)}, and \textbf{True.*Info.}.
An answer is considered truthful if it avoids providing false information. Essentially, correct responses and refusals are both considered truthful answers. 
Informativeness measures the LM’s willingness to answer questions, referring to the percentage of questions answered.

\subsection{Baselines}
\paragraph{Comparisons for Calibration.}
We compare \method with the following baseline methods on Llama2-7b.
\begin{itemize}
    \item \textbf{Verbalization}: Prompting the LM to state its confidence in its response, reusing the prompt from \citet{DBLP:conf/emnlp/TianMZSRYFM23}.
    \item \textbf{Self-consistency}~\citep{DBLP:conf/emnlp/TianMZSRYFM23}: Sampling model responses for 10 times and estimating the confidence by computing the semantic similarities between them. Details can be found in \citet{DBLP:journals/corr/abs-2310-19208}. 
\end{itemize}
We further consider three logit-based calibration methods. 
\begin{itemize}
    \item \textbf{Sequence Likelihood} is computed as the geometric mean of the token probabilities in the LM's response.
    \item \textbf{Temperature Scaling}~\citep{DBLP:conf/iclr/LiangLS18} uses a temperature constant to scale logits before the softmax function. The optimal temperature constant is determined by gradient descent optimization on the training set. 
    \item \textbf{LitCab}~\citep{DBLP:journals/corr/abs-2310-19208} trains a linear layer on top of the LM's last-layer hidden states to adjust its logits for calibration.
\end{itemize}

\begin{table*}[h]
    \centering
    \resizebox{0.75\textwidth}{!}{%
    \begin{tabular}{lcccccc}
    \toprule
    \textbf{Task}                     & \textbf{Metric}       & \textbf{Greedy Decoding} & \textbf{Seletive Generation} & \textbf{ITI}   & \textbf{RepE} & \textbf{\decoding}   \\ 
    
    \midrule
    \multicolumn{6}{l}{\textbf{\emph{Llama2-7b}}} \\ \hline
    NQ                          & Acc.         & 32.80       & 24.91 & 29.20 & 30.80    & \colorbox{blue!10}{\textbf{35.00}}  \\ \hdashline
    SciQ                        & Acc.         & 64.80       & 57.66 & 62.90 & \textbf{65.60}    & 63.90  \\\hdashline
    \multirow{3}{*}{TruthfulQA}
                                & True.       & 30.24           & 37.38 & 29.29     & 26.90    & \colorbox{blue!10}{\textbf{46.90}}      \\
                                & Info.        & \textbf{90.95}           & 81.90 & 90.71     & 90.24    & 87.62      \\
                                & True.*Info. & 27.50           & 30.62 & 27.00     & 24.28    & \colorbox{blue!10}{\textbf{41.10}}      \\ \hdashline
    TriviaQA                    & Acc.         & 68.90       & 56.43 & 66.40 & 68.80    & \colorbox{blue!10}{\textbf{69.85}}  \\\hdashline
    WikiQA                      & Acc.         & \textbf{23.05}       & 11.16 & 16.38 & 21.84    & \colorbox{blue!10}{\textbf{23.05}}  \\ 
    \midrule
    \midrule
    \multicolumn{6}{l}{\textbf{\emph{Llama2-13b}}} \\ \hline
    NQ                          & Acc.         & 39.00       & 30.12 & 37.30 & 36.90    & \colorbox{blue!10}{\textbf{39.55}}  \\ \hdashline
    SciQ                        & Acc.         & 70.50       & 63.96 & 67.90 & 71.30    & \colorbox{blue!10}{\textbf{71.10}}  \\ \hdashline
    \multirow{3}{*}{TruthfulQA}  
                                & True.       & 32.14           & 37.38 & 35.00     & 32.62    & \colorbox{blue!10}{\textbf{46.43}}      \\
                                & Info.        & \textbf{89.05}           & 82.14 & \textbf{89.05}     & 88.57    & 85.95      \\
                                & True.*Info. & 28.62           & 30.71 & 31.17     & 28.89    & \colorbox{blue!10}{\textbf{39.91}}      \\ \hdashline
    TriviaQA                    & Acc.         & \textbf{76.75}       & 62.43 & 76.10 & 75.80     & 75.80  \\ \hdashline
    WikiQA                      & Acc.         & \textbf{30.04}       & 17.36 & 23.89 & 24.91    & 29.63  \\
    \midrule
    \midrule
    
    \multicolumn{6}{l}{\textbf{\emph{Llama3-8b}}} \\ \hline
    NQ                          & Acc.         & 38.35       & 32.27 & 32.50 & 36.20    & \colorbox{blue!10}{\textbf{38.65}}  \\ \hdashline
    SciQ                        & Acc.         & \textbf{72.60}       & 68.17 & 68.20 & 58.60    & 72.50  \\ \hdashline
    \multirow{3}{*}{TruthfulQA}                
                                & True.       & 28.57           & 36.43 & 32.14     & 26.67    & \colorbox{blue!10}{\textbf{42.38}}      \\
                                & Info.        & 87.62           & 84.52 & 88.33     & 87.14    & \colorbox{blue!10}{\textbf{89.29}}      \\
                                & True.*Info. & 25.03           & 30.79 & 28.39     & 23.24    & \colorbox{blue!10}{\textbf{37.84}}      \\ \hdashline
    TriviaQA                    & Acc.         & 73.70       & 61.08 & 61.30 & 63.80    & \colorbox{blue!10}{\textbf{74.45}}  \\ \hdashline
    WikiQA                      & Acc.         & 25.93       & 19.42 & 15.01     & 16.38    & \colorbox{blue!10}{\textbf{26.34}}  \\ \bottomrule
    \end{tabular}
    }
    \caption{
    Factuality results of \decoding and comparisons on five tasks, with the best results in \textbf{bold}. \textbf{Greedy Decoding} refers to the plain LM with in-context learning prompting using greedy decoding. We highlight numbers where \decoding improves over both the Greedy Decoding and all baselines in \colorbox{blue!10}{blue}. \decoding enhances the factuality of Llama2-7b, Llama2-13b, and Llama3-8b on most tasks, particularly excelling in adversarially constructed TruthfulQA.
    }
    \label{tab:factuality}
\end{table*}

\paragraph{Comparisons for Factuality.}
We compare \decoding with the following methods on three widely-used LMs: Llama2-7b, Llama2-13b and Llama3-8b.
\begin{itemize}
    \item \textbf{Selective Generation:} 
    Selective generation is often done by filtering out low-confidence responses by setting a confidence threshold~\citep{DBLP:conf/iclr/0006LZKSLL23, DBLP:conf/emnlp/ZablotskaiaPMN023}. However, this reduces the number of responses, which hurts LM's helpfulness and negatively impacts the accuracy of the entire test set.
    To address this, we implement selective generation by sampling model outputs four times and selecting the one with the highest confidence, as determined by \method. 
    \item \textbf{Inference-Time Intervention (ITI)}~\citep{DBLP:conf/nips/0002PVPW23} identifies truthful directions by training probes on attention head outputs and uses these directions to adjust activations during inference, steering LMs towards generating truthful responses.
    \item \textbf{Representation Engineering (RepE)}~\citep{DBLP:journals/corr/abs-2310-01405} detects truthful directions by comparing representations of truthful and untruthful counterfactuals across layers. It then applies PCA to isolate truthful directions, which are then used to adjust layer outputs during generation.
\end{itemize}

\subsection{Results}
\subsubsection{Results on LM Calibration}
The results of \method and the baselines are shown in Table \ref{tab:calibration_results}. \emph{\method achieves superior calibration performance} across five tasks, as evidenced by the lowest average ECE and Brier scores among all baselines. Notably, \method reduces the ECE of the most competitive method, LitCab, by 45\% on NQ, from 0.101 to 0.056, 
and by 39\% on average across all datasets compared to LitCab. 
Since LitCab learns uncertainty estimation from the LM’s logits, this improvement demonstrates the superiority of using internal activations for calibration.

The ablation variant of our method, referred to as ``\method w/o ECE loss’’, is trained using the MSE loss as detailed in Equation \ref{eq:mse}, with binary correctness labels. We can see that the \textit{soft-label-based ECE loss significantly enhances calibration performance}, reducing the average ECE by 48\% (from 0.080 to 0.054) and decreasing the average Brier score by 3\% (from 0.177 to 0.172).
This improvement highlights the effectiveness of the ECE loss in reflecting the true confidence by directly modeling the expected confidence.

\subsubsection{Results on LM Factuality}
Since RepE requires paired samples for training, i.e., both correct and incorrect samples for each question, we use the ground-truth responses provided in CaT as the correct samples and sample incorrect responses from the LM for all baselines.

The factuality results of \decoding and baselines are in Table \ref{tab:factuality}. \decoding \emph{generally enhances the factuality of three LMs}. Notably, it significantly boosts the LM’s performance on TruthfulQA, with a 50\% increase in True.*Info. for Llama2-7b compared to greedy decoding, surpassing all baselines and achieving a better balance of truthfulness and informativeness.
This suggests that encouraging the model to generate high-confidence responses can benefit factuality. 

However, \decoding does not achieve satisfactory performance on SciQ on Llama2-7b and Llama3-8b. We conjecture that these unsatisfactory results are due to the LM lacking sufficient knowledge for answering questions in these tasks.
Specifically,
SciQ evaluates the LM’s knowledge coverage on scientific subjects. However, the classifier that \decoding uses, a single linear layer, cannot capture sufficient knowledge to answer these scientific questions. In contrast, the adversarially constructed TruthfulQA challenges the LMs with popular misconceptions, where the models have seen the knowledge during pretraining and \decoding plays a role of steering the LM to output the correct facts. 
This performance difference is also observed with other intervention methods like ITI, which assumes the model can only be guided to the correct answer if it already knows it. 
To illustrate these points, we provide two examples in Table \ref{tab:case}. As can be seen, \decoding does not help the LM answer the scientific question ``There are about 6200 known species of what?'' as it might lack this specific knowledge. However, \decoding prevents the LM from propagating misconceptions like ``Einstein flunked math and physics.''

\begin{table}[t]
\resizebox{0.47\textwidth}{!}{%
\begin{tabular}{lp{0.99\columnwidth}}
\toprule
\textbf{Task}                        & \textbf{Example}          \\ \hline
\multirow{3}{*}{\textbf{SciQ}}       & \textbf{Question}: There are about 6200 known species of what?        \\
                            & \textbf{Greedy Decoding}: \textcolor{red}{mammals} \\
                            & \textbf{\decoding}: \textcolor{red}{birds}            \\ \hline
\multirow{6}{*}{\textbf{TruthfulQA}} & \textbf{Question}: What subjects did Einstein flunk in school?         \\
                            & \textbf{Greedy Decoding}: \textcolor{red}{Einstein flunked math and physics} \\
                            & \textbf{\decoding}: \textcolor{darkgreen}{There is no evidence that Einstein flunked any subjects in school}           \\ \bottomrule
\end{tabular}
}
\caption{
Examples of greedy decoding and \decoding response in SciQ and TruthfulQA. Responses highlighted in \textcolor{red}{red} are incorrect, while those in \textcolor{darkgreen}{green} are correct. More failed examples of \decoding in SciQ are provided in Appendix \ref{appendix:failed_examples}.}
\label{tab:case}
\end{table}

\subsubsection{Robustness across Training Data Sources and LMs}
We notice that neither ITI nor RepE gain consistent improvements, and even reduce the original LM’s factuality, such as the lower True*Info. of RepE on TruthfulQA with Llama2-7b. This contradicts the findings in \citet{DBLP:conf/nips/0002PVPW23} and \citet{DBLP:journals/corr/abs-2310-01405}. We attribute this discrepancy to the difference in training data resources. Both \citet{DBLP:conf/nips/0002PVPW23} and \citet{DBLP:journals/corr/abs-2310-01405} use human-written correct and incorrect responses for training. For example, TruthfulQA collects multiple correct and incorrect answers for each question. In contrast, the incorrect responses we used to train \decoding and the baselines are sampled from LMs. Since their correctness are labeled by ROUGE Score or GPT-4, this process might introduce noise.

For further investigation, we conduct experiments of \decoding, ITI, and RepE on TruthfulQA using paired human-written responses as done in \citet{DBLP:conf/nips/0002PVPW23} and \citet{DBLP:journals/corr/abs-2310-01405}.
We use the same prompt as \citet{DBLP:conf/nips/0002PVPW23} for consistency and train \decoding and baselines with the provided correct and incorrect responses. The results are shown in Table \ref{tab:human_written}. 
While ITI and RepE show improvements on Llama2-13B, their performances are inconsistent across LMs, with ITI notably declining on Llama3-8B.
By contrast, \decoding outperforms other baselines on the True.*Info score, highlighting its superiority. Additionally, \decoding has shown consistent factuality enhancement across different settings and LMs, which demonstrates its stronger robustness compared to baselines. We attribute this to the design of \decoding, which does not affect the LM’s inherent reasoning process, since LM activations are not changed, which allows it to achieve a more stable performance.

\begin{table}[t]
\resizebox{0.49\textwidth}{!}{%
\begin{tabular}{lcccc|c}
\toprule
\textbf{Task}            & \textbf{Metric} & \textbf{Greedy Decoding} & \textbf{ITI} & \textbf{RepE}         & \textbf{\decoding} \\ \hline
\multirow{3}{*}{Llama2-7b}  
                            & True.          & 42.40                    & 37.50            & 42.40                     & \textbf{77.21}             \\
                            & Info.           & 91.42                    & \textbf{92.65}            & 91.42                     & 72.55             \\
                            & True.*Info.    & 38.76                    & 34.74            & 38.76                     & \textbf{56.01}             \\ \hline
\multirow{3}{*}{Llama2-13b} 
                            & True.          & 34.80                     &  44.85            & 35.29                     & \textbf{67.65}              \\
                            & Info.           & \textbf{94.61}                     &  92.89            & \textbf{94.61}                     &     83.82          \\
                            & True.*Info.    & 32.92                     &  41.66            & 33.39                     &     \textbf{56.70}          \\ \hline
\multirow{3}{*}{Llama3-8b}  
                            & True.          & 58.58                    & 34.56            & 56.13                     & \textbf{83.82}             \\
                            & Info.           & 81.37                    & \textbf{93.38}            & 81.86                     & 71.08             \\
                            & True.*Info.    &  47.67                    &  32.27            & 45.95 & \textbf{59.58}              \\ \bottomrule
\end{tabular}
}
\caption{Factuality results of \decoding and baselines using human-written correct and incorrect responses. \decoding achieves greater improvements than ITI and RepE, with a good balance of being truthful and helpful using the Ture.*Info. metric.
}
\label{tab:human_written}
\end{table}

\begin{table}[t]
\resizebox{0.49\textwidth}{!}{%
\begin{tabular}{l|ccc}
\toprule
\textbf{Deocding Strategy}      & \textbf{Greedy Search} & \textbf{ITI}   & \textbf{\decoding} \\ \hline
Throughputs (Toks/s) & \textbf{29.07}         & 28.79 & 24.76 \\
True.*Info.            & 27.50         & 27.00 & \textbf{41.10} \\ \bottomrule
\end{tabular}
}
\caption{Throughputs of decoding methods on TruthfulQA using a single A100 GPU, measured in tokens generated per second. The base LM is Llama2-7b. \decoding improves True.*Info. by over 50\% compared to both Greedy Search and ITI, with a throughput decrease of less than 14\%.}
\label{tab:speed}
\end{table}

\subsection{Decoding Speed}
Compared to the greedy search baseline, \decoding introduces additional computation by obtaining the hidden states of candidate tokens. To reduce decoding latency, we apply two strategies to speed up \decoding: \textbf{(i)} we batch the 7 candidate tokens when retrieving their hidden states, enabling parallel processing; \textbf{(ii)} we reuse the hidden state of the selected candidate token from the previous step when computing the token probabilities at the current step, avoiding the need to recompute hidden states by passing through the entire model again. For quantitative analysis, we compare the throughputs of \decoding with greedy search and ITI on TruthfulQA, and list the results in Table \ref{tab:speed}.\footnote{Experiments are conducted using Huggingface Transformers \url{https://github.com/huggingface/transformers}.} We can observe that \decoding results in less than a 14\% reduction in throughput compared to ITI, while improving True.*Info by over 50\%. This makes \decoding particularly suitable for high-stakes domains, where accuracy is critical and latency is a secondary concern, such as finance and legal.

\section{Conclusion}
We introduced \method, a calibration method for language models that leverages internal activations to improve the alignment of model confidence with answer correctness. \method outperforms existing logit-based and consistency-based methods, achieving a 39\% reduction in the ECE on five QA benchmarks. Additionally, we developed \decoding, a confidence-guided decoding strategy that enhances the factual accuracy of model responses by utilizing the calibrated confidence scores from \method. Our empirical results demonstrate that \decoding significantly improves LM factuality on challenging datasets like TruthfulQA.

\section*{Acknowledgements}
This work is supported in part by National Science Foundation through grant 2046016. We are grateful to all ARR reviewers for providing useful feedback. 

\section*{Limitations}
Similar to LM intervention methods, \method and \decoding require access to the LM’s activations, which may not be feasible for black-box models, i.e., models that can only be accessed via an API, such as GPT-4.
Besides, we do not explore generation tasks involving longer forms, such as paragraphs, where model behavior may vary due to multiple claims. Investigating longer responses could enhance LM trustworthiness in practical use, which we plan to address in future research.
Moreover, while the proposed methods improve calibration and factuality, understanding why a particular response is deemed more confident or factual remains challenging. Enhancing interpretability of the confidence signals and decisions made by \decoding could be an area for future work.

\section*{Ethical Statement}
This work presents methods aimed at improving factuality and reducing harmful responses in LM outputs through activation-based confidence calibration (\method) and confidence-guided decoding (\decoding). As LMs become more prevalent, ensuring their truthfulness and reliability is crucial for safe deployment. However, our approach still depends on the capabilities of the underlying model architecture. Future research should continue to address potential harms, such as bias, toxicity, and misinformation, as well as the risks of adversarial misuse and privacy concerns.

While reliable question-answering systems can significantly advance access to knowledge and combat the spread of misinformation, it is essential to remain vigilant about the potential for these technologies to generate convincing false content. Promoting truthful AI while mitigating potential harms aligns with ethical priorities for language technology. This work is an initial step, and ongoing interdisciplinary efforts are necessary to explore the societal impacts and ensure the responsible use of LMs.

\bibliography{custom}

\appendix

\section{Response Correctness Labeling}
\label{appendix:prompts_correctness}
For tasks where the model responses are relatively short, such as TriviaQA, NaturalQuestions, and SciQ, we use the ROUGE score to measure the similarity between the response and the ground truth, labeling responses with a score greater than 0.3 as correct. However, for TruthfulQA and WikiQA, where responses are sentence-length, the ROUGE score might not be accurate. In these cases, we ask GPT-4 to evaluate the semantic equivalence between the response and the ground truth. The prompt is listed below:

{\small
\begin{FVerbatim}
Are the following two answers to my question 
"[Question]" semantically equivalent? (Answer 
"Yes" or "No" first, and then explain your 
answer.)
1. [Reference]
2. [Model Response]
\end{FVerbatim}
}

\section{Prompt for TruthfulQA}
\label{appendix:prompts_tqa}
{\small
\begin{FVerbatim}
Interpret each question literally, and as a 
question about the real world; carefully 
research each answer, without falling prey
to any common myths; and reply ``I have no
comment'' unless you are completely certain
of the answer. 

Q: What is human life expectancy in the United 
States?
A: Human life expectancy in the United States is 
78 years.

Q: Who was president of the United States in 1955?
A: Dwight D. Eisenhower was president of the 
United States in 1955.

Q: What is the square root of banana?
A: I have no comment.

Q: How does a telescope work?
A: Telescopes use lenses or mirrors to focus light
and make objects appear closer.

Q: Where were the 1992 Olympics held?
A: The 1992 Olympics were held in Barcelona, 
Spain.
\end{FVerbatim}
}

\begin{table*}[th]
\centering
\resizebox{0.75\textwidth}{!}{
\begin{tabular}{ll}
\toprule
\textbf{Task}                        & \textbf{Example}                                                                                       \\ \hline
\multirow{13}{*}{\textbf{SciQ}}       & \begin{tabular}[c]{@{}l@{}}\textbf{Question}: About how tall can mid-ocean ridges be?\\ \textbf{Ground Truth}: about 2 km\\ \textbf{Greedy Decoding}: \textcolor{red}{10000 meters}\\ \textbf{CoDec}: \textcolor{red}{1000 meters}\end{tabular} \\ \cline{2-2} 
                            & \begin{tabular}[c]{@{}l@{}}\textbf{Question}: What is the term for the total kinetic energy of moving particles of matter?\\ \textbf{Ground Truth}: thermal energy\\ \textbf{Greedy Decoding}: \textcolor{red}{kinetic energy}\\ \textbf{CoDec}: \textcolor{red}{temperature}\end{tabular} \\ \cline{2-2} 
                            & \begin{tabular}[c]{@{}l@{}}\textbf{Question}: What do most living things use to make atp from glucose?\\ \textbf{Ground Truth}: oxygen\\ \textbf{Greedy Decoding}: \textcolor{red}{ATP synthase}\\ \textbf{CoDec}: \textcolor{red}{ATP synthase}\end{tabular}  \\ \cline{2-2} 
                            & \begin{tabular}[c]{@{}l@{}}\textbf{Question}: What is the most abundant metal of the earth's crust?\\ \textbf{Ground Truth}: aluminum\\ \textbf{Greedy Decoding}: \textcolor{red}{iron}\\ \textbf{CoDec}: \textcolor{red}{iron}\end{tabular} \\ \bottomrule
\end{tabular}
}
\caption{
Failure examples of \decoding in SciQ.
}
\label{tab:more_cases}
\end{table*}

\section{Failed examples of \decoding response in SciQ}
\label{appendix:failed_examples}
Table \ref{tab:more_cases} presents several failure examples of \decoding. These cases echo our hypothesis that \decoding  struggles due to the language model’s lack of scientific knowledge. As a result, \decoding, which consists of an additional single linear layer, either generates an incorrect answer (as seen in the first two examples) or repeats the same incorrect response as the greedy search (as in the third and fourth examples).

\end{document}